\documentclass[conference]{IEEEtran}
\IEEEoverridecommandlockouts
\usepackage{cite}
\usepackage{amsmath,amssymb,amsfonts}
\usepackage{algorithmic}
\usepackage{graphicx}
\usepackage{textcomp}

\usepackage{color}
\usepackage{url}
\usepackage{graphicx}
\usepackage{subfigure}
\usepackage{multirow}
\usepackage{booktabs}
\usepackage{lineno}
\usepackage{amsmath}
\usepackage{graphicx}
\usepackage[ruled]{algorithm2e}

\usepackage{newfloat}
\usepackage{listings}
\usepackage{mathtools}
\usepackage{bbm}
\usepackage{amsfonts,amssymb}
\usepackage{enumitem}
\usepackage[T1]{fontenc}
\usepackage{cleveref}
\usepackage{wrapfig}
\usepackage{lineno}
\newcommand\figcaption{\def\@captype{figure}\caption}
\usepackage{adjustbox}
\usepackage[table]{xcolor}

\makeatletter
\newcommand{\linebreakand}{%
  \end{@IEEEauthorhalign}
  \hfill\mbox{}\par
  \mbox{}\hfill\begin{@IEEEauthorhalign}
}
\makeatother

\def\BibTeX{{\rm B\kern-.05em{\sc i\kern-.025em b}\kern-.08em
    T\kern-.1667em\lower.7ex\hbox{E}\kern-.125emX}}
\begin{document}

\title{Learning Cautiously in Federated Learning \\with Noisy and Heterogeneous Clients
}



\author{Chenrui Wu$^{1,2,*}$, Zexi Li$^{5,*}$, Fangxin Wang$^{2,1,3,4,\dag}$, Chao Wu$^{6,\dag}$ \\
$^1$The Future Network of Intelligence Institute, The Chinese University of Hong Kong, Shenzhen \\
$^2$School of Science and Engineering, The Chinese University of Hong Kong, Shenzhen\\

$^3$The Guangdong Provincial Key Laboratory of Future Networks of Intelligence\\
$^4$Peng Cheng Laboratory\\
$^5$ College of Computer Science and Technology, Zhejiang University\\
$^6$ School of Public Affairs, Zhejiang University\\
Email: chenruiwu@link.cuhk.edu.cn, zexi.li@zju.edu.cn, wangfangxin@cuhk.edu.cn, chao.wu@zju.edu.cn
\thanks{$^*$Equal contributions. $^\dag$Corresponding authors.}
\thanks{The work is supported in part by the Basic Research Project No. HZQB-KCZYZ-2021067 of Hetao Shenzhen-HK S\&T Cooperation Zone, by National Natural Science Foundation of China (Grant No. 62102342), by Guangdong Basic and Applied Basic Research Foundation (Grant No. 2023A1515012668), by Shenzhen Science and Technology Program (Grant No. RCBS20221008093120047), by Shenzhen Outstanding Talents Training Fund 202002, by Guangdong Research Projects No. 2017ZT07X152 and No. 2019CX01X104, by the Guangdong Provincial Key Laboratory of Future Networks of Intelligence (Grant No. 2022B1212010001), by Young Elite Scientists Sponsorship Program by CAST (Grant No. 2022QNRC001) and by The Major Key Project of PCL Department of Broadband Communication, by the National Key Research and Development Project of China (2021ZD0110400), National Natural Science Foundation of China (U19B2042), Program of Zhejiang Province Science and Technology (2022C01044), The University Synergy Innovation Program of Anhui Province (GXXT-2021-004), Academy Of Social Governance Zhejiang University, Fundamental Research Funds for the Central Universities (226-2022-00064).}

}
\maketitle

\begin{abstract}
Federated learning (FL) is a distributed framework for collaborative training with privacy guarantees. In real-world scenarios, clients may have Non-IID data (local class imbalance) with poor annotation quality (label noise). The co-existence of label noise and class imbalance in FL’s small local datasets renders conventional FL methods and noisy-label learning methods both ineffective. To address the challenges, we propose \textsc{FedCNI} without using an additional clean proxy dataset. It includes a noise-resilient local solver and a robust global aggregator. For the local solver, we design a more robust prototypical noise detector to distinguish noisy samples. Further to reduce the negative impact brought by the noisy samples, we devise a curriculum pseudo labeling method and a denoise Mixup training strategy. For the global aggregator, we propose a switching re-weighted aggregation method tailored to different learning periods. Extensive experiments demonstrate our method can substantially outperform state-of-the-art solutions in mix-heterogeneous FL environments. 
\end{abstract}


\section{Introduction}

The proliferation of smart devices such as mobile phones, cameras, and sensors has dramatically expanded the service of multimedia applications like live broadcasts and real-time video analytics~\cite{zhang2022towards}. The massive data plays a key role in generating powerful predictive models to provide better services to users. However, transferring users' data to the server poses a high privacy risk and brings huge communication burdens, rendering traditional centralized training ineffective. Therefore, Federated Learning (FL) \cite{pmlr-v54-mcmahan17a,fedprox,feddyn} stands out as a promising solution that enables such collaborative training only by aggregating local models uploaded from user clients without any data exchange. 

\begin{figure}[t]
\centering
\subfigure[Small-loss on majority] 
{\includegraphics[width=3.4cm]{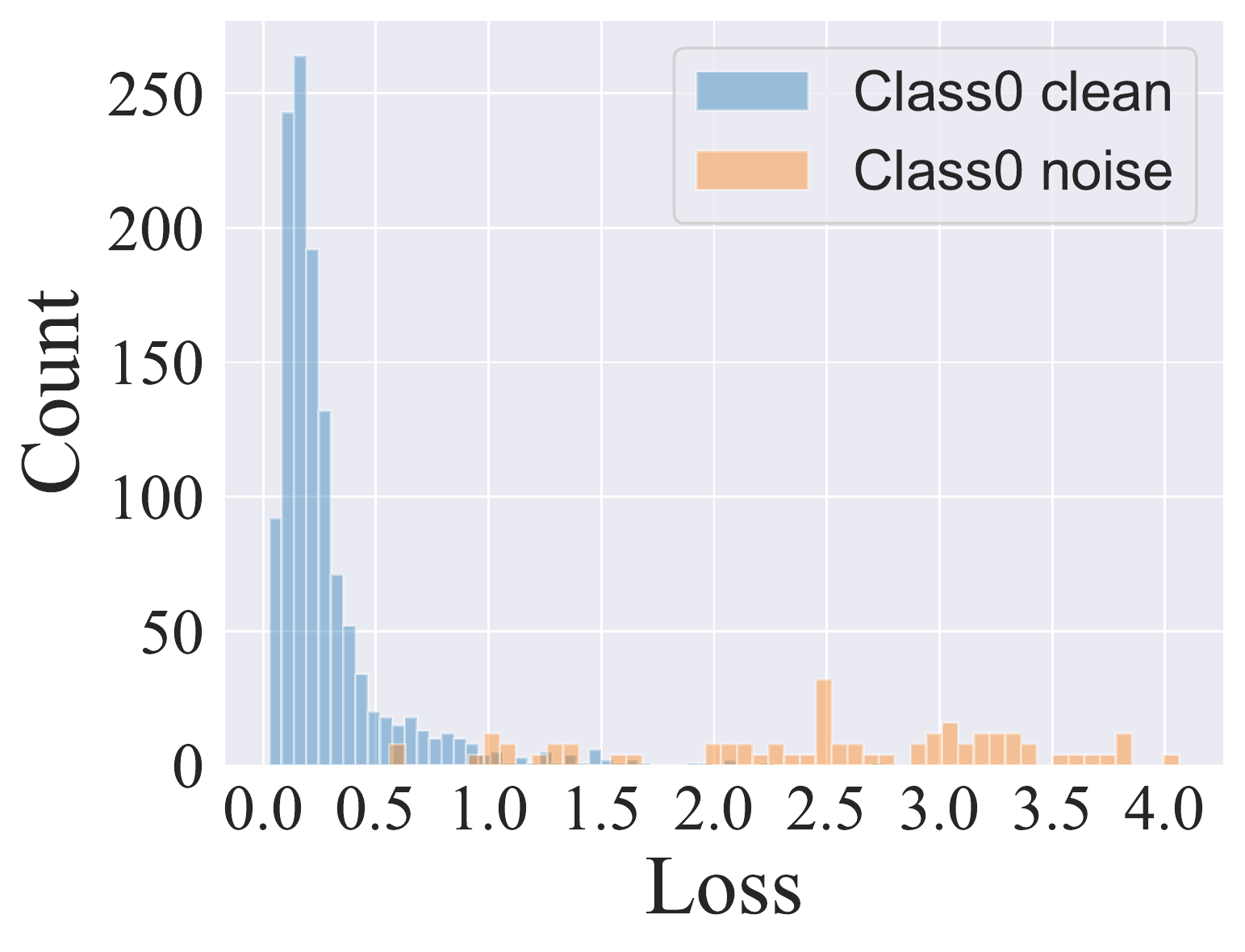}} 
\subfigure[Small-loss on minority]{\includegraphics[width=3.4cm]{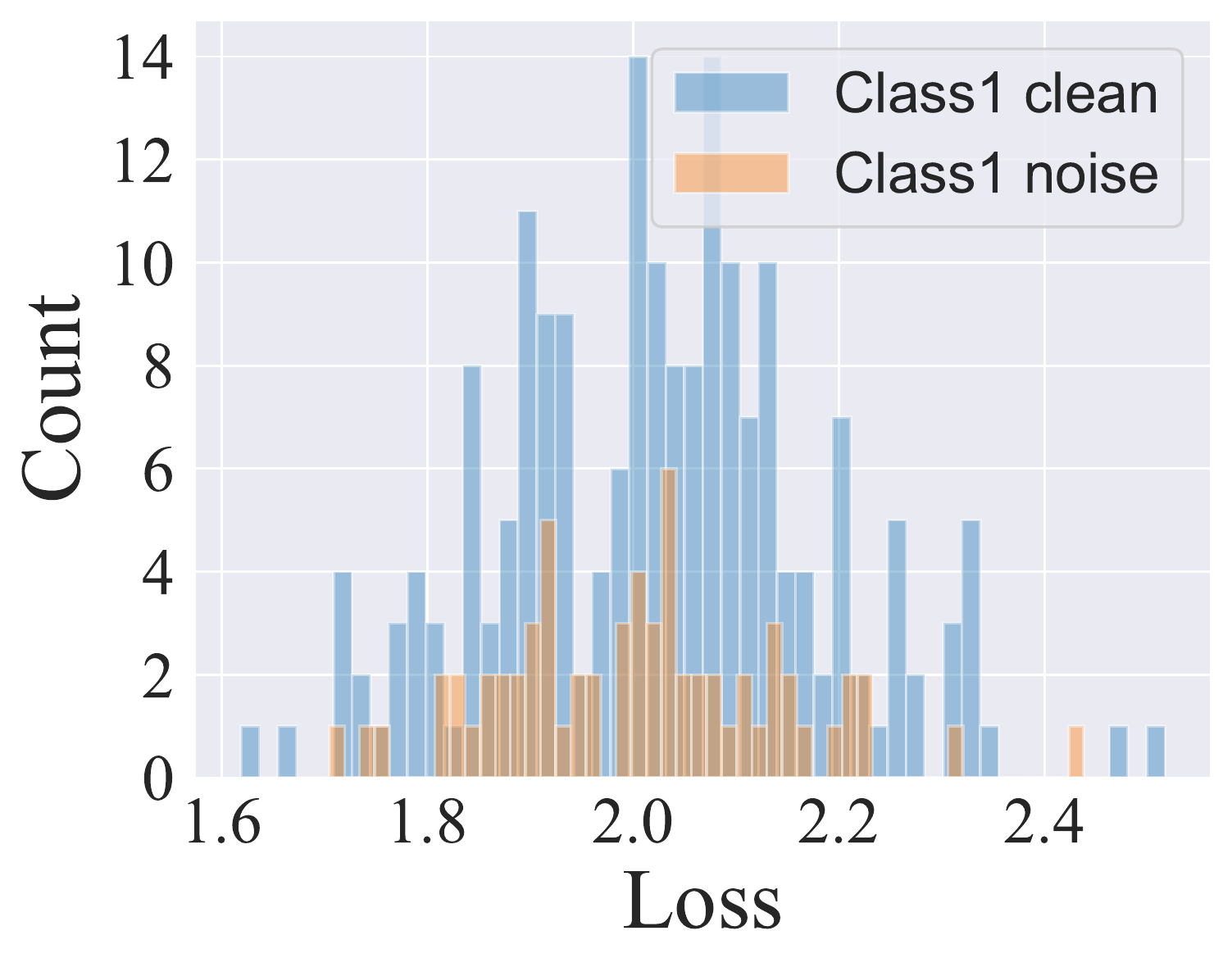}}
\subfigure[Our on majority]{\includegraphics[width=3.4cm]{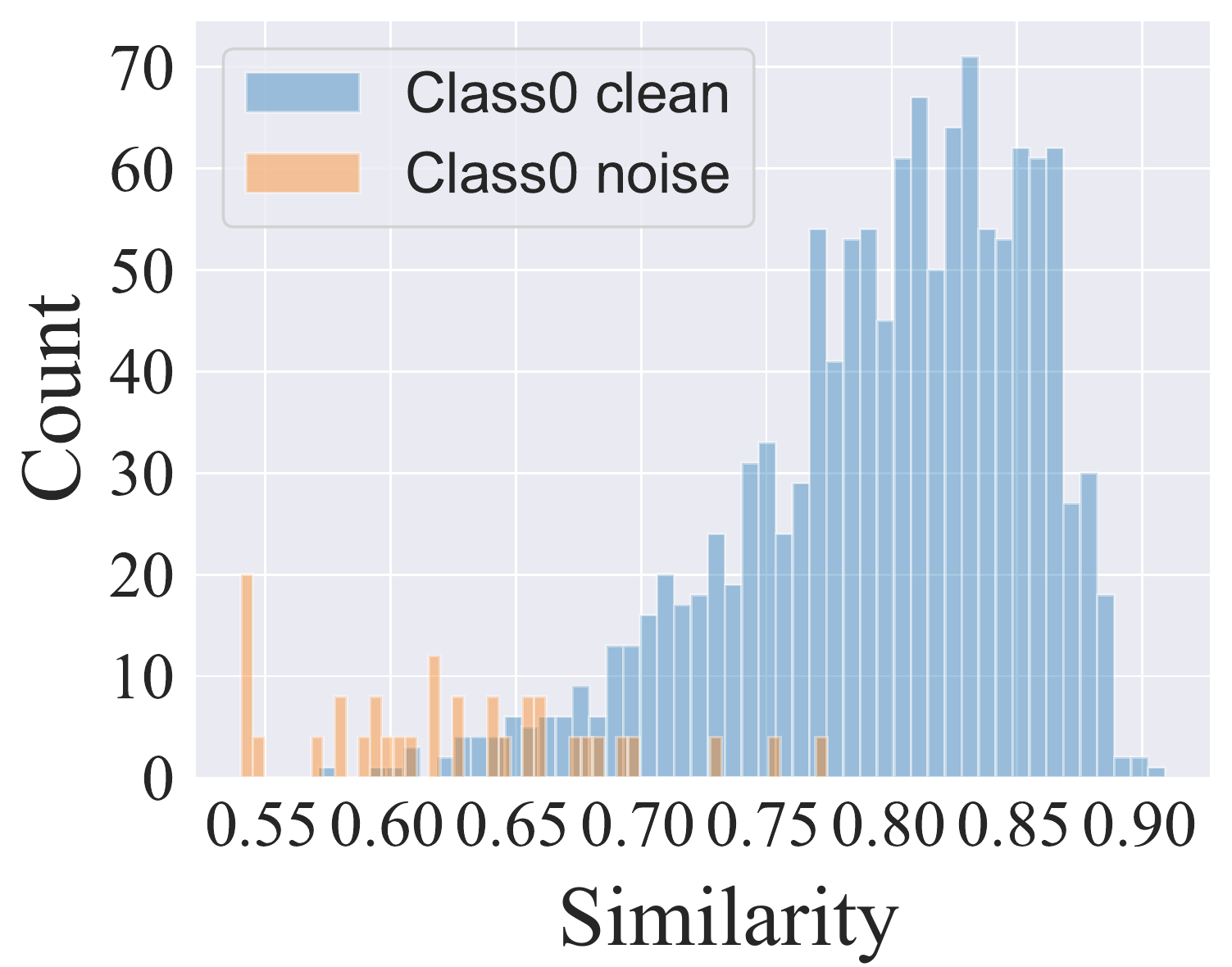}}
\subfigure[Our on minority]{\includegraphics[width=3.4cm]{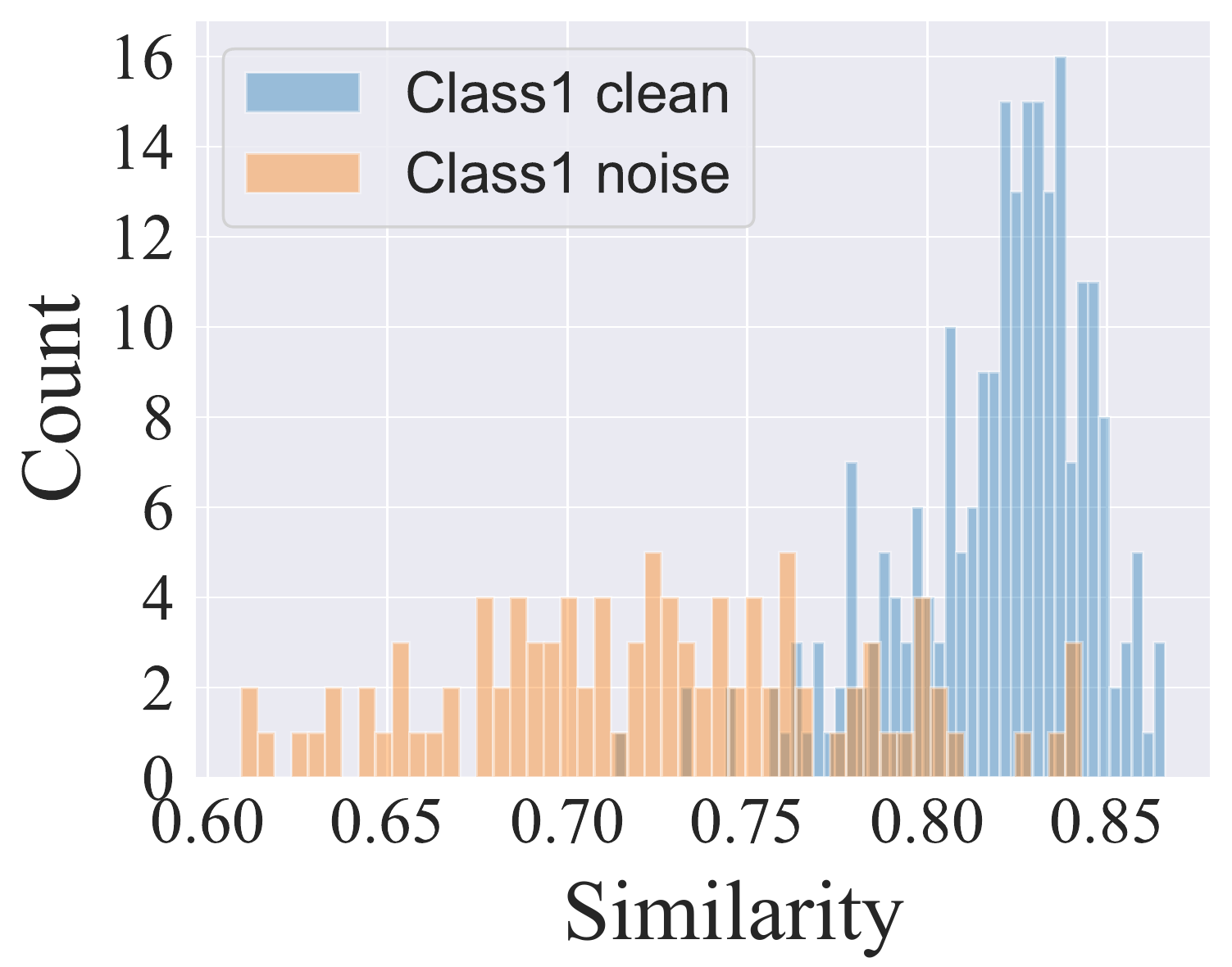}}
\vspace{-10pt}
\caption{ \textbf{Illustration of noise detection results on FL's small and imbalanced local datasets.} Class 0 is a majority class and Class 1 is a minority class. (a) and (b) show that the small-loss method can distinguish noisy samples in the majority class but fails in the minority class. However, in (c) and (d), our proposed prototypical method is effective in both the majority and minority.}
\label{distribution figure}  
 \vspace{-15pt}
\end{figure}
Data heterogeneity is an inherent problem in FL, as clients may have Non-IID data, especially for class distributions. Previous works have explored this issue from both local and global perspectives \cite{fedprox,9860009}. Besides, poor annotation quality is also a naturally arising problem in FL. In practical FL scenarios, labels of local data are often machine-generated or manually annotated. Nevertheless, clients have different domain expertise and various human biases in annotation, resulting in inaccurate labeling and heterogeneous label noise among clients \cite{Fang_2022_CVPR,xu2022fedcorr,9713942}. 

The joint problem of local class imbalance and label noise is very challenging. On the one hand, the existence of label noise makes conventional FL methods, which mainly tackle Non-IID data, ineffective (cf. Table~\ref{cifa10100}). On the other hand, conventional noisy-label learning methods used in centralized training have poor performance in FL due to the class imbalance and small sizes of local datasets (cf. Table~\ref{cifa10100}). One intuitive example is shown in Figure~\ref{distribution figure}, that the small-loss technique \cite{li2019dividemix,NEURIPS2018_a19744e2}, which is commonly used in noise detection of centralized learning, no longer works when the client's data has imbalanced classes, noisy labels, and small sizes. The small dataset makes the model poorly generalized overall, and the classifier learned from class-imbalanced data has even worse generalization on the minority classes. As a result, both clean and noisy samples of the minority class have large losses and low confidence, which is hard to distinguish\footnote{We note that the poor local noise detection will impede the local model performance and further result in a poor global model after aggregation.}. Moreover, clients have diverse imbalances and noise levels, and it is also challenging for the central server to aggregate.

There are some pioneer works addressing the problem of FL with noisy labels. However, they are not tailored to solve the noisy-label learning problem jointly with local class imbalance. For the noise detection methods, they mainly rely on the small-loss technique \cite{xu2022fedcorr,9713942}, which is shown to be less effective. Besides, the existing methods hold strong assumptions or preconditions that are less realistic in practical situations. Some works rely on clean proxy datasets on the server \cite{9412599,chen2020focus}, clean public datasets held by all clients \cite{Fang_2022_CVPR}, or clean clients without noisy labels \cite{xu2022fedcorr}. However, clean datasets are infeasible since collecting the clients' data are forbidden in FL, and annotating such a clean dataset requires huge costs. Additionally, the clean client assumption is not satisfied if the FL system is rather heterogeneous. 

Thus, in this paper, we propose \textbf{Fed}erated \textbf{C}autious learning for \textbf{N}oisy and \textbf{I}mbalanced clients (\textsc{\textbf{FedCNI}}) to cautiously learn from the noisy and highly-skewed data in FL without using an additional clean proxy dataset. \textsc{FedCNI} cautiously learns by accurately detecting noise (with both high precision and recall) and reducing the negative impact of the detected noisy samples. It includes a noise-resilient local solver and a robust global aggregator. For the local solver, we first design a prototypical noise detector to distinguish noisy samples, and it is shown to be more robust than the small-loss-based noise detector. To reduce the negative impact brought by the noisy samples, we devise a curriculum pseudo labeling method to dynamically assign labels to the detected noisy samples and propose a denoise Mixup training strategy to separate the clean and noisy sets in Mixup. For the global aggregator, we propose a switching re-weighted aggregation method tailored to different learning periods. Extensive experiments under both synthetic and natural label noise demonstrate our method can outperform state-of-the-art solutions in mix-heterogeneous FL environments by a large margin. In some cases, our method even has comparable performance with \textsc{FedAvg} under purely clean data. Our contributions can be concluded as follows:
\begin{itemize}[leftmargin=*]
    \item We propose the \textsc{FedCNI} tailored for tackling the joint challenge of label noise and class-imbalanced data in real-world FL scenarios, which is a pioneering work in FL.
    \item In \textsc{FedCNI}, we craft a noise-resilient local solver and a robust global aggregator without resorting to clean proxy datasets or clean client assumptions, which have robust performance in mix-heterogeneous FL environments.
    \item We conduct extensive experiments to show that \textsc{FedCNI} outperforms state-of-the-art FL methods over multiple datasets under both synthetic and natural label noise.
\end{itemize}

\begin{figure*}[!h]
    \centering
    \includegraphics[width=2.0\columnwidth]{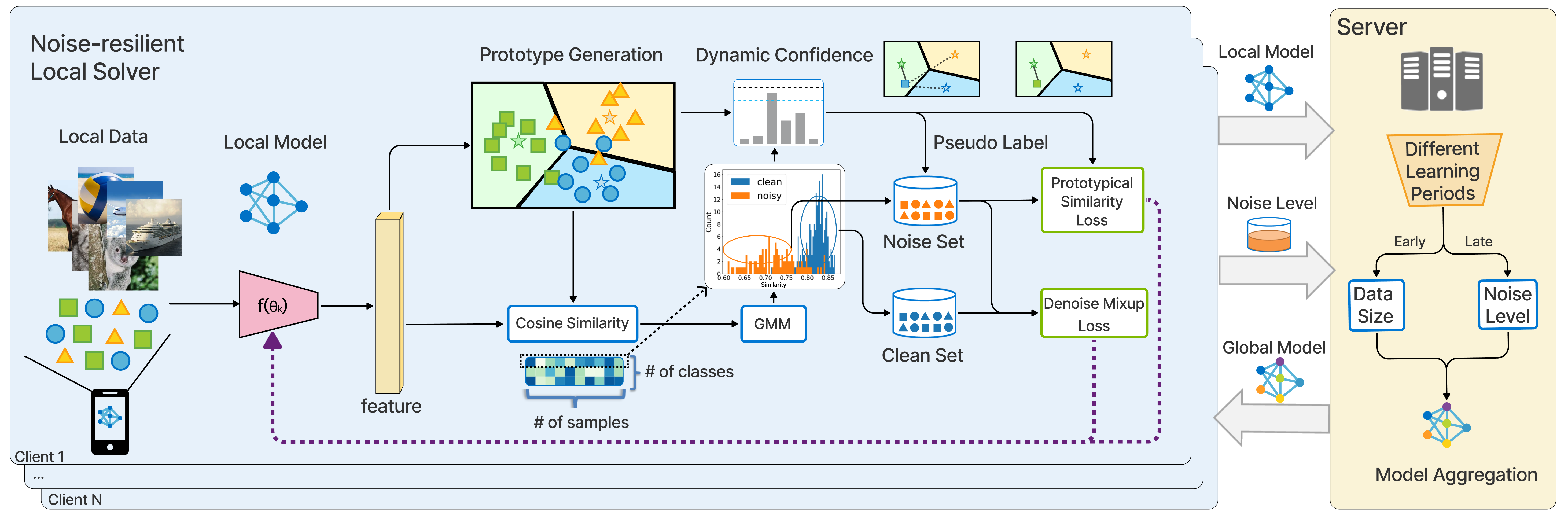}
    \vspace{-10pt}
    \caption{\textbf{The framework overview of the proposed \textsc{FEDCNI}.}}
    \label{framwork}
      \vspace{-10pt}
\end{figure*} 
\section{Related works}
\label{related work}
\noindent\textbf{Heterogeneous Data in Federated Learning.} Clients have Non-IID data distributions is an inherent problem in FL, which is also known as data heterogeneity. 
There are a lot of works in FL that focus on tackling this. 
\textsc{FedProx} \cite{fedprox} is proposed with a proximal term on the client side, so the model parameters obtained by the client after local training will not deviate too much from the initial server parameters. \textsc{FedDyn} \cite{feddyn} adds a regularization term in local training based on the global model and the model from previous rounds of communication to overcome device heterogeneity. 

\noindent\textbf{Noisy-label Federated Learning.} There are some previous works that have concerned the FL with noisy labels. \textsc{FedCorr} \cite{xu2022fedcorr} is a multi-stage FL algorithm that includes noise client detection, noisy sample detection and correction, and a vanilla \textsc{FedAvg}~\cite{pmlr-v54-mcmahan17a} phase.  However, the method's multi-step LID score calculations require high computational complexity and extensive hyper-parameter tuning. \textsc{RoFL} \cite{9713942} introduces the exchange of class centroids between the server and clients to give pseudo labels and generates loss based on similarity, which may threaten clients' privacy due to the direct transfer of class centroids. Additionally, the small-loss technique utilized by \textsc{FedCorr} \cite{xu2022fedcorr} and \textsc{RoFL} \cite{9713942} performs poorly in class-imbalance scenarios and requires clean clients, which is not practical in real-world applications. There are also some existing works that rely on a clean proxy (benchmark) dataset on the server side. In \cite{9412599}, they upload local samples' loss distribution to the server for noise detection by proxy dataset. The transmitted loss distribution can raise severe privacy concerns. \textsc{RHFL} \cite{Fang_2022_CVPR} interchanges model logits and analyze them by a public dataset. The method lacks noise detection, only depending on KL divergence of knowledge. Besides, the auxiliary public dataset is not available for the server and clients in real applications. 
\section{Methodology} \label{method}
In this section, we will first describe the problem formulation with some basic notations and then introduce \textsc{FedCNI}.
\subsection{Problem Formulation}
\label{problem formulation}
We consider a typical FL scenario with a multi-class classification task. There are $K$ clients and overall $N$ data samples in training. Each client $k\in\{1,\dots,K\}$ holds a private dataset $\mathcal{D}_k= \{ (x_i^k,y_i^k)\}_{i=1}^{n_k}$, where $x_i^k$ is the input of the training sample, corresponding $y_i^k$ denotes the given label, and the number of local samples is ${n_k}$ ($\sum_{k=1}^K n_k = N$). In the inaccurate annotation scenario, $y_i^k  \in \{1,\dots,C\}$ can be the same as the ground-truth label $\bar{y}_i^k$, or be different as a noise. We use $\theta$ to denote the model parameter, and for client $k$, it has the local model $\theta_k$. We use $\mathbb{F}$ to denote the feature extraction function of a network, given a model $\theta$ and a sample $x$, the extracted feature embedding is $\mathbb{F}(\theta;x)$. In addition, we use $\mathbb{P}$ to denote the prediction function (the last layer), given a model $\theta$ and a sample $x$. The predicted softmax is $\mathbb{P}(\theta;x)$.

\subsection{\textsc{FedCNI} Overview}
Our method \textsc{FedCNI} consists of a noise-resilient
local solver and a robust global aggregator. For the noise-resilient
local solver, it has two modules: the prototypical local noise detection (cf. Section \ref{noise detection}) and the noise-resilient local loss (cf. Section \ref{local loss}). The prototypical local noise detection distinguishes noisy samples via prototypical similarity and assigns labels to these noisy samples by curriculum pseudo labeling with dynamic confidence. The noise-resilient local loss includes a prototypical similarity loss and a denoise Mixup loss, and it facilitates the clients to cautiously learn from the rare and noisy local data. For the global aggregator (cf. Section \ref{global aggregator}), a switching re-weighted aggregation method tailored to different learning periods is proposed. The framework overview is shown in Figure~\ref{framwork}.

\subsection{Prototypical Local Noise Detection}
\label{noise detection}

\noindent\textbf{Prototype Generation.} The class prototype $p_c$ is defined as a normalized mean of samples' embeddings for the class $c \in \{1, \dots C\}$. For client $k$, we can obtain its local prototypes as:
\begin{equation}
\label{protoeqation}
   p_{k,c}= \frac{1}{|\mathcal{D}_{k,c}|} \sum_{(x_i^k,y_i^k)\in \mathcal{D}_{k,c} } \mathbb{F}(\theta_k;x_i^k),
\end{equation}
where $\mathcal{D}_{k,c}$ denotes the samples given the label $c$ in local dataset $\mathcal{D}_k$ and $\mathbb{F}(\theta_k;x_i^k)$ refers to the output embedding of sample $x_i^k$ given the model $\theta_k$. 


\noindent\textbf{Noise Detection.} 
For client $k$, given a prototype of a class $p_{k,c}$, we compute the embeddings of the samples which are labeled as $c$. Intuitively, the embeddings of clean samples may have high similarities to the prototype, while the noisy samples represent outliers in the embedding space. Thus, we use a two-component Gaussian Mixture Model (GMM) in the similarities to distinguish noisy and clean samples for each class. 
We adopt the cosine function as the similarity measurement, which is effective in high dimensional space. The cosine similarity between a sample $x_i^k \in \mathcal{D}_k$ and a prototype $p_{k,c}$ is given by:
\begin{equation}
   \cos(p_{k,c},x_i^k)= \frac{\langle p_{k,c},\mathbb{F}(\theta_k;x_i^k) \rangle}{\Vert p_{k,c}\Vert\cdot \Vert \mathbb{F}(\theta_k;x_i^k)\Vert}.
\end{equation}
We can obtain class-wise similarity sets $s_{k,c} = \{\cos(p_{k,c},x_i^k)| x_i^k \in \mathcal{D}_{k,c}\}$ that contain the similarities between the class prototype and the samples. We conduct local noise detection using a two-component GMM in $s_{k,c}$ and obtain the noisy set with lower similarity $\mathcal{N}_{k,c}$ and clean set with higher similarity $\mathcal{C}_{k,c}$. Then the overall detected noisy set for client $k$ is $\mathcal{N}_k = \sum_{c=1}^{C}\mathcal{N}_{k,c}$, and the clean set is $\mathcal{C}_k = \sum_{c=1}^{C}\mathcal{C}_{k,c}$. The noisy and clean sets are disjoint such that $\mathcal{N}_k \cap \mathcal{C}_k = \emptyset, \mathcal{N}_k \cup \mathcal{C}_k = \mathcal{D}_k$.

\noindent\textbf{Curriculum Pseudo Labeling.} 
Given the detected noisy samples, we use the pseudo labeling method to correct the labels.
However, due to different noise and imbalance levels of clients, the difficulty of pseudo labeling in each class of each client is different. 
Inspired by the idea of curriculum labeling in semi-supervised learning \cite{zhang2021flexmatch}, we propose a novel curriculum pseudo labeling method providing dynamic confidence for labeling threshold. 

We define the dynamic threshold $\tau_{k,c}$ for each class $c$ in client $k$. Every class $c \in \{1, \dots, C\}$ has a different level of noise and data size; therefore, $\tau_{k,c}$ is to describe the difficulty of learning this class. Let $q_i^k=\max_{j}\mathbb{P}^j(\theta_k;x_i^k)$ and $\hat{y}_i^k=\arg\max_{j}(\mathbb{P}^j(\theta_k;x_i^k))$, where $\mathbb{P}(\theta_k;x_i^k)$ is the prediction softmax values and $\mathbb{P}^j(\theta_k;x_i^k)$ is the $j$-th value of the softmax. We introduce the definition of learning difficulty $\rho_{k,c}$ as:
\begin{equation}
\label{learning effect}
    \rho_{k,c}=\frac{ \sum_{(x_i^k,y_i^k)\in \mathcal{C}_{k,c}}     \mathbb{I}(q_i^k > \tau_{k,c}) \cdot    \mathbb{I}(\hat{y}_i^k=c)    }{|\mathcal{D}_{k,c}|},
\end{equation}
In Equation \ref{learning effect}, the initial threshold $\tau_{k,c}$ is set as $\tau$ for all classes after the early phase of training. 
In the equation, we quantify the learning difficulty $\rho_{k,c}$ as a confident and clean sample proportion for each class. Then, we normalize $\rho_{k,c}$ to update the dynamic threshold $\tau_{k,c}$, as:
\begin{equation}
\label{confidence}
    \tau_{k,c}=\frac{\rho_{k,c}}{\max(\rho_{k})}\tau, \text{ where } \rho_{k} = \{\rho_{k,c}| c \in \{1, \dots, C\}\}.
\end{equation}
We update $\rho_{k,c}$ and $\tau_{k,c}$ iteratively as in Equation \ref{learning effect} and Equation \ref{confidence} during local training. 

Besides, we can also get the cosine similarities between one sample $x_i^k$ and all classes' prototypes $\{\cos(p_{k,c},\mathbb{F}(\theta_k;x_i^k))\}_{c=1}^C$, and we use these cosine similarities as the prototypical classifier to predict the pseudo labels for the detected noisy set $\mathcal{N}_k$. Concretely, given a sample $(x_i^k,y_i^k) \in \mathcal{N}_k$, the pseudo label is as
\begin{equation}
    \widetilde{y}_i^k = \arg\max_{c}(\{\cos(p_{k,c},\mathbb{F}(\theta_k;x_i^k))| c \in \{1, \dots, C\}\}).
\end{equation}

We can also derive a max value of the softmax cosine similarities, denoted as: $ \widetilde{q}_i^k =  {\rm softmax  }( \{\cos(p_{k,c},\mathbb{F}(\theta_k;x_i^k))\}_{c=1}^{C})$, showing the confidence of pseudo labeling. According to the dynamic thresholds, if the confidence value of the pseudo label is higher than the threshold, we assign the sample with the pseudo label, otherwise, we use its original label, as
\begin{equation} \label{equ:pseudo_label}
y_i^k = \begin{cases}
\widetilde{y}_i^k ,& {\rm if  }\; \widetilde{q}_i^k > \tau_{k,\widetilde{y}_i^k}, \\
y_i^k, & {\rm otherwise }. \\
\end{cases}
\end{equation}
\subsection{Noise-resilient Local Loss}
\label{local loss}
Given the detected noisy samples and the corresponding pseudo label in Section \ref{noise detection}, we now devise the noise-resilient local loss to cautiously learn from the noisy data by treating the detected clean and noisy samples differently. The loss consists of two parts, the first is the denoise Mixup loss and the second is the prototypical similarity loss.

\noindent\textbf{Denoise Mixup Loss.} Recall that Mixup \cite{zhang2018Mixup} is a data augmentation method that mixes up the samples' features and labels to generate new samples. Specifically, it generates new sample $(\tilde{x},\tilde{y})$ by linear combination of randomly selected pairs of samples $(x_{i}, y_{i})$ and $(x_{j}, y_{j})$, as $\tilde{x}=\lambda x_i+(1-\lambda)x_j, \tilde{y}=\lambda y_i+(1-\lambda)y_j$. Mixup has been shown to be effective in both semi-supervised learning \cite{zhang2021flexmatch} and noisy-label learning \cite{li2019dividemix}. But we notice it has marginal gains in noisy-label FL. We think this is because the clients' local data are rare and the samples with wrong labels will have larger negative effects in the Mixup process. Intuitively, randomly mixing up the wrong-label samples with other samples will generate more noisy-label samples. 
Therefore, we propose a denoise Mixup loss to treat the noisy samples $\mathcal{N}_k$ and clean samples $\mathcal{C}_k$ differently. 
Specifically, for the detected noisy samples, we mix them with the samples from their corresponding class ${y}_i^k$ (in Equation \ref{equ:pseudo_label}) to reduce the effects of wrong labels, while for the clean data, we adopt vanilla Mixup (i.e. randomly-mixed) within the clean set. Given a sample $(x,y)$, we use $(\tilde{x},\tilde{y})$ to denote the corresponding-class-only Mixup strategy and $(\hat{x},\hat{y})$ to denote the randomly-mixed Mixup strategy. Thus, our denoise Mixup loss for client $k$ is formulated as 
\begin{equation}
\label{Mixuploss}
 \begin{split}
\mathcal{L}_{k}^{mix}(\theta_k)=&\frac{1}{| \mathcal{N}_k |}\sum_{(x_i^k,y_i^k) \in \mathcal{N}_k} \mathcal{L}_{mix}(\theta_k;(\tilde{x}_i^k,\tilde{y}_i^k))\\
&+ \frac{1}{| \mathcal{C}_k |}\sum_{(x_j^k,y_j^k) \in \mathcal{C}_k} \mathcal{L}_{mix}(\theta_k;(\hat{x}_j^k,\hat{y}_j^k)).
\end{split}
\end{equation}

\noindent\textbf{Prototypical Similarity Loss.} Moreover, we consider the similarity of a noisy sample and its pseudo label's prototype as also a learning point. According to experimental evaluations, we observe that the pseudo label precision is confident enough to reduce the gap between a noisy sample and its corresponding prototype. Thus, we devise a prototypical similarity loss for the noisy samples $\mathcal{N}_k$.
\begin{equation}
\label{distanceloss}
\mathcal{L}_{k}^{sim}(\theta_k)=\frac{1}{| \mathcal{N}_k |}\sum_{(x_i^k,y_i^k) \in \mathcal{N}_k}(1 - \cos(p_{k,{y}_i^k},\mathbb{F}(\theta_k;x_i^k))).
\end{equation}
Overall, the noise-resilient loss of client $k$ is the sum of the two mentioned losses, formulated as:
\begin{equation} \label{eq:sum_loss}
\mathcal{L}_k^{sum}(\theta_k)=\mathcal{L}_k^{mix}(\theta_k)+\lambda_{sim}\mathcal{L}_k^{sim}(\theta_k),
\end{equation}
where $\lambda_{sim}$ is a hyper-parameter controlling the strength of $\mathcal{L}_k^{sim}$. Clients locally adopt SGD for $E$ epochs to minimize $\mathcal{L}_k^{sum}$ and then send the updated weights to the server. 

\subsection{Robust Global Aggregator}
\label{global aggregator}
Learning with label noise has different training dynamics in learning periods \cite{pmlr-v97-shen19e}, it is found that in the early period, generalization takes place that the neural networks learn the correct samples which have common patterns, while in the late, the networks memorize the noisy data and fail in generalization. Inspired by this observation, we find that applying different re-weighted aggregations in the early and late will improve the generalization. Thus, we propose a switching aggregation strategy. We denote $T_{s}$ as the switching round for aggregation; during the early period (before round $T_{s}$), where mostly the generalization takes place, we adopt the data-size-based aggregation as \textsc{FedAvg}; while in the late period (after round $T_{s}$), where bad memorization may occur, we cautiously aggregate clients' models according to the noise levels. Hence, the robust global aggregator is as follows, where $N=\sum_{k=1}^{K} n_{k}$ is the sum of local data sizes, and $M=\sum_{k=1}^{K} |\mathcal{C}_k|$ is the sum of local clean data sizes. 
\begin{equation}
\label{aggregation}
\theta^{t+1} = \begin{cases}
\sum_{k=1}^{K} \frac{|\mathcal{D}_k|}{N}   \theta_k^t, & t < T_{s}, \\
\sum_{k=1}^{K} \frac{|\mathcal{C}_k|}{M}   \theta_k^t ,& t \geq T_{s}. \\
\end{cases}    
\end{equation}

\setlength{\tabcolsep}{2pt}
\begin{table*}[ht!]
\vspace{-1cm}
\centering
\caption{\textbf{Test accuracies (Top-1\% with 5$\times$-Avg ) on CIFAR-10/100 datasets with symmetric and pair flipping label noise.}}
\label{cifa10100}
\resizebox{\textwidth}{!}{
\begin{tabular}{l|cc|cc|cc|cc|cc|cc|cc|cc}
\toprule 
\multicolumn{1}{c|}{\multirow{4}{*}{Method}} & \multicolumn{8}{c|}{CIFAR-10}  & \multicolumn{8}{c}{CIFAR-100 }    \\ \cline{2-17} 
\multicolumn{1}{c|}{}  & \multicolumn{4}{c|}{Symmetric} & \multicolumn{4}{c|}{Pair}  & \multicolumn{4}{c|}{Symmetric} & \multicolumn{4}{c}{Pair}   \\ \cline{2-17} 
\multicolumn{1}{c|}{}                        & \multicolumn{2}{c|}{$\mu=0.2$, $\sigma=0.2$} & \multicolumn{2}{c|}{$\mu=0.4$, $\sigma=0.2$} & \multicolumn{2}{c|}{$\mu=0.2$, $\sigma=0.2$} & \multicolumn{2}{c|}{$\mu=0.4$ , $\sigma=0.2$}  & \multicolumn{2}{c|}{$\mu=0.2$, $\sigma=0.2$} & \multicolumn{2}{c|}{$\mu=0.4$, $\sigma=0.2$} & \multicolumn{2}{c|}{$\mu=0.2$, $\sigma=0.2$} & \multicolumn{2}{c}{$\mu=0.4$ , $\sigma=0.2$}       \\ \cline{2-17} 

\multicolumn{1}{c|}{}                        & $\alpha$=1 & \multicolumn{1}{c|}{$\alpha$=0.7} & $\alpha$=1 & \multicolumn{1}{c|}{$\alpha$=0.7} & $\alpha$=1 & \multicolumn{1}{c|}{$\alpha$=0.7} & $\alpha$=1           & $\alpha$=0.7  & $\alpha$=1 & \multicolumn{1}{c|}{$\alpha$=0.7} & $\alpha$=1 & \multicolumn{1}{c|}{$\alpha$=0.7} & $\alpha$=1 & \multicolumn{1}{c|}{$\alpha$=0.7} & $\alpha$=1           & $\alpha$=0.7           \\ 
\midrule
\textsc{FedAvg}  & 79.94 & 77.95 & 64.89 & 63.28 
                & 80.58&81.10&67.98& 60.42 
                & 48.61  & 47.62 & 36.35& 37.63  & 52.54  & 53.18 & 38.43 & 37.62            \\
\textsc{FedProx }     & 79.25 &77.75 & 66.21 & 64.12
                      & 79.63&80.46&62.30&63.54
                      &48.2  & 47.58 & 37.22 & 36.6 & 52.93 & 52.24 & 39.26 & 38.13\\
\textsc{FedDyn}  &76.56 & 70.22 & 60.54 & 60.07
                 & 70.11 & 69.07 & 55.36&57.88
                 & 1.08 & 1.2 & 1.36 & 1.21 & 1.14 & 1.31 &1.1 & 1.28 \\
\textsc{SCAFOLLD }  & 79.15& 77.87 &63.79 & 66.29 
                    & 80.29& 79.26&61.12&61.84&49.45  & 47.73 & 36.79 & 36.46 & 54.01 & 51.16 & 38.92 &  39.01          \\
\midrule
\textsc{FedCorr} &79.02&78.18  & 64.02 & 61.16
                    &76.24 &78.47&55.72 &63.07 
                    &37.74 & 39.02 &37.09 &38.53 &52.77 &50.07& 40.21 &39.91      \\
\textsc{RoFL }   &80.59 & 81.71 &65.27 & 65.92
                &  81.37& 80.22 & \textbf{72.21 }& 65.32
                & 47.71 &48.09 & 45.5 & 48.09 &47.15 &46.49 &44.83 & 46.72      \\
\textsc{FedProto  }  &82.11& 80.87  &  72.16 &  71.93 
                    &80.09 &  82.36 & 61.67 &   69.74
                    &  55.83 &  55.2  &45.84 & 44.78&  59.78 &  58.11& 45.29 &  44.28       \\
\midrule
Distributed \textsc{Co-teaching}  & 31.72 & 29.76 & 28.30 &26.28  &28.65  &26.43&22.33 &21.14 
& 9.71&7.98 &   8.42 &8.15&8.19 &8.75&7.20& 6.55            \\
Distributed \textsc{DivideMix}  & 48.81 & 44.09 &34.58 &31.14&46.64 &43.57 & 33.46&33.02 
&17.60& 18.57& 14.52&13.45&16.34&18.24&14.16&13.92

\\
\textsc{FedAVG+Co-teaching}  &45.95 &29.78 &39.85&32.63
                &42.93& 35.73 & 30.43& 22.14 &16.09 &14.37 & 9.47 &8.48 & 15.73& 13.7 &9.73  &8.52      \\
\textsc{FedAVG+DivideMix   }  &80.23& 79.68 & 65.17 & 60.72
                    &74.15 & 76.13&54.31&54.19 
                    &39.28  &37.18 &30.06 & 28.69& 45.25&44.37  &35.12 &27.05    \\

\midrule
\rowcolor{gray!20}Ours   & \textbf{ 86.62}  & \textbf{ 84.38 }  &\textbf{  78.02 }& \textbf{ 78.45} 
        &\textbf{86.13}& \textbf{ 82.37} & 72.16 &\textbf{  71.04} & \textbf{ 62.13} &\textbf{  56.42 }& \textbf{ 54.37 }& \textbf{ 50.29} &\textbf{61.33 } & \textbf{ 59.14}  & \textbf{ 53.07}& \textbf{ 50.42}\\
 \bottomrule
\end{tabular}
}
\vspace{-10pt}
\end{table*}
\begin{table*}[ht!] 
\caption{\textbf{Top-1\% test accuracy on Clothing1M with natural label noise. }}
  \centering
  \label{clothing}
  \begin{adjustbox}{width=1.7\columnwidth,center}
  \begin{tabular}{l|cccccccccc}
    \toprule
    Dataset/ Method  & \textsc{FedAvg} & \textsc{FedProx} &\textsc{FedDyn} &\textsc{SCAFFOLD} &\textsc{FedCorr} &\textsc{RoFL} &\textsc{FedProto} &  \textsc{Co-teaching} & \textsc{DivideMix} &Ours\\
    \midrule
    Clothing1M  &68.34 &69.85 &70.55 & 69.36& 72.4&73.31 &70.52 &69.83&70.1&\textbf{74.26}
    \\
    \bottomrule
  \end{tabular}
  \end{adjustbox}
  \label{table: clothing_noniid}
  \vspace{-10pt}
\end{table*}
\section{Experiments}
\label{experiment}
\subsection{Experimental Setting}
\noindent\textbf{Datasets and Models.}
We follow the existing works\cite{xu2022fedcorr,9713942}, and apply Resnet-18~\cite{he2016deep} for CIFAR-10\cite{krizhevsky2009learning} , Resnet-34 for CIFAR-100, and Resnet-50 for Clothing1M\cite{Xiao_2015_CVPR}. We note that Clothing1M is a dataset with natural label noise, and for  CIFAR-10/100, we corrupt them via synthetic label noise.

\noindent\textbf{Data Partition and Noise Distribution.} (i) We adopt a general Non-IID data partition in FL by Dirichlet distribution \cite{9860009,tan2022fedproto,9713942}, where $\alpha$ controls the heterogeneity, the smaller, the more Non-IID. We consider a more Non-IID setting than previous works in noisy-label FL, where $\alpha \in \{1,~0.7\}$. (ii) To generate real-world noisy labels in heterogeneous FL environments, we allocate a truncated Gaussian distribution to formulate the noise level for each client. We sample two groups of noise levels, a lower group is $\mu=0.2,\sigma=0.2$, and a higher noise level group is $\mu=0.4,\sigma=0.2$. We corrupt CIFAR-10/100 with two widely-used types of label noise: symmetric
flipping \cite{9713942,ghosh2017robust} and pair flipping \cite{NEURIPS2018_a19744e2}.

\noindent\textbf{Baselines.} We compare \textsc{FedCNI} with the following state-of-the-art methods in four groups: i) general FL methods: \textsc{FedAvg} \cite{pmlr-v54-mcmahan17a}, \textsc{FedProx} \cite{fedprox}, \textsc{FedDyn} \cite{feddyn}, and \textsc{SCAFFOLD} \cite{scaffold}; ii) a prototype based FL solution: \textsc{FedProto} \cite{tan2022fedproto}; iii) methods designed for label noise in centralized learning: \textsc{Co-teaching} \cite{NEURIPS2018_a19744e2} and \textsc{DivideMix} \cite{li2019dividemix}, and we construct a distributed implementation and a combination with \textsc{FedAvg}; iv) FL methods to tackle label noise without proxy datasets: \textsc{FedCorr} \cite{xu2022fedcorr} and \textsc{RoFL}\cite{9713942}. For other methods requiring a clean proxy dataset, like RHFL \cite{Fang_2022_CVPR}, it is not fair for comparison, so we exclude them from baselines.

\noindent\textbf{Implementation Details.} We use 20 clients fully participating in FL training in each round. We use the SGD optimizer with a learning rate of 0.01 and momentum of 0.5. The entire FL training process will last for 100 rounds to ensure convergence. We keep the number of local epochs $E=5$ and the local batch size as 100 in all experiments. For the switching round, we set $T_s=15$. The default confidence is set as $\tau =0.5$, the hyper-parameter $\lambda_{sim} =0.7$. 


\subsection{Main Results}
\noindent\textbf{Synthetic Label Noise.} We compare \textsc{FedCNI} with state-of-the-art methods in multiple noise types, noise levels, and imbalance levels on CIFAR-10/100 datasets, shown in Table~\ref{cifa10100}. Generally, \textsc{FedCNI} achieves the best test accuracy under both symmetric-flip or pair-flip noise types, and its advantage is dominant at high noise levels. For CIFAR-10, \textsc{FedCNI} consistently outperforms all baselines except \textsc{FedProto} by at least 5\%, while in CIFAR-100, our method demonstrates superior performance with performance gains ranging from 2\% to 9\%. 
The conventional FL methods are not robust in noisy-label environments that are marginal above \textsc{FedAvg} (or even worse). 
We also compare the noisy-label FL baselines, and due to the extremely imbalanced local data, they are not robust and effective enough. It is worth noting that \textsc{FedProto}, which aggregates updates based on prototypes, consistently surpasses the other baselines. It indicates that the prototype-based methods are more robust in mix-heterogeneous FL environments, and we further take this advantage into noise detection and training loss in our method. 
Additionally, we simply combine centralized noisy-label learning methods as local solvers and conduct \textsc{FedAvg}. It is shown that the \textsc{DivideMix} variants have better performances than the \textsc{Co-teaching} variants, but they all perform worse than the FL baselines. It proves that the simple combination cannot address FL's inherent challenges, which further supports our contribution. 


\noindent\textbf{Natural Label Noise.} Table~\ref{clothing} shows the results on the real-world dataset Clothing1M with natural label noise. It is obvious that \textsc{FedCNI} also has the best performance.
\vspace{-0.3em}

\begin{figure*}[t]

 \vspace{-1cm}
\centering
\subfigure[Test accuracy curve] 
{\includegraphics[width=0.62\columnwidth]{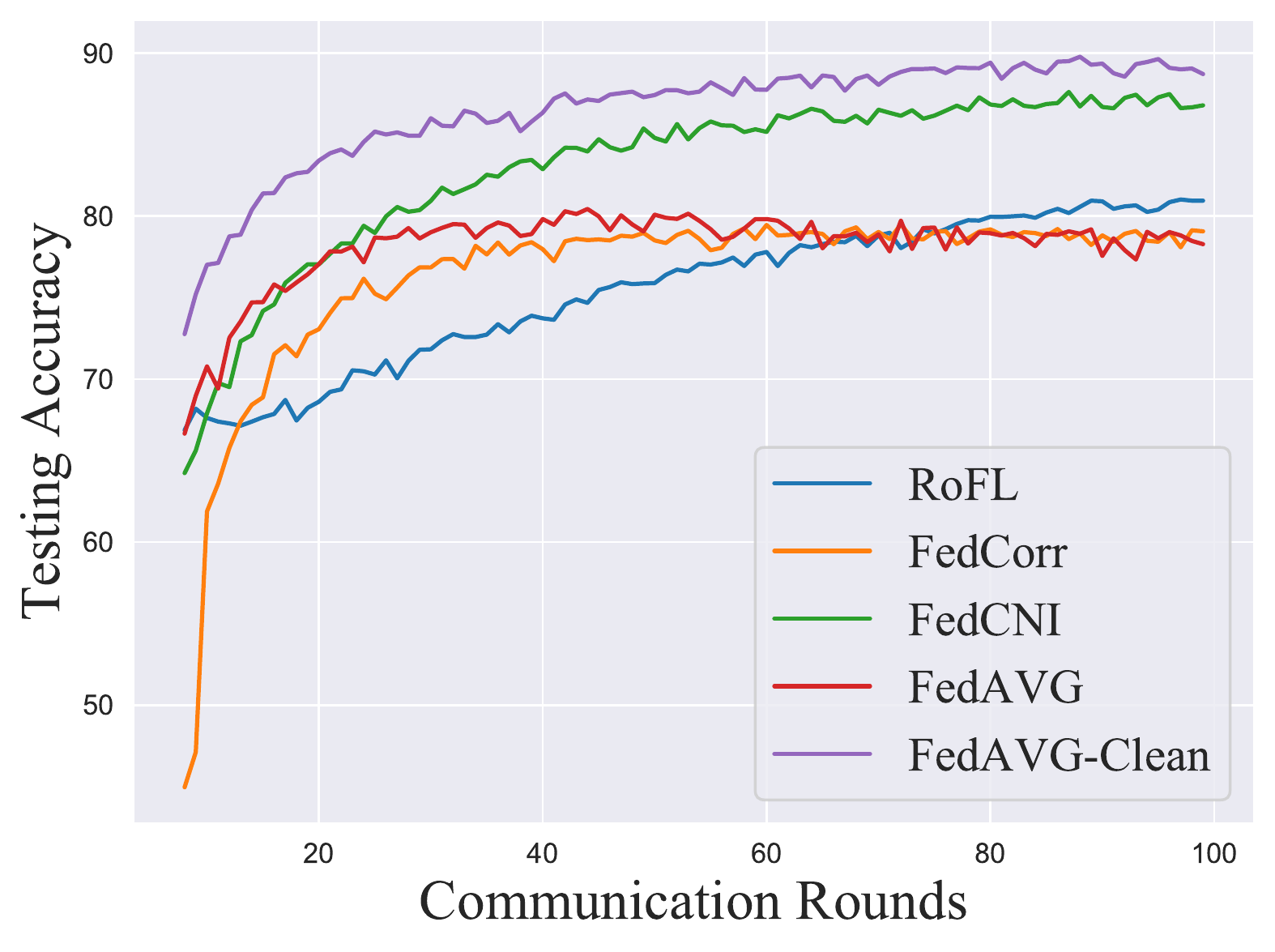}} 
\subfigure[Precision and recall of noise detection]{\includegraphics[width=0.62\columnwidth]{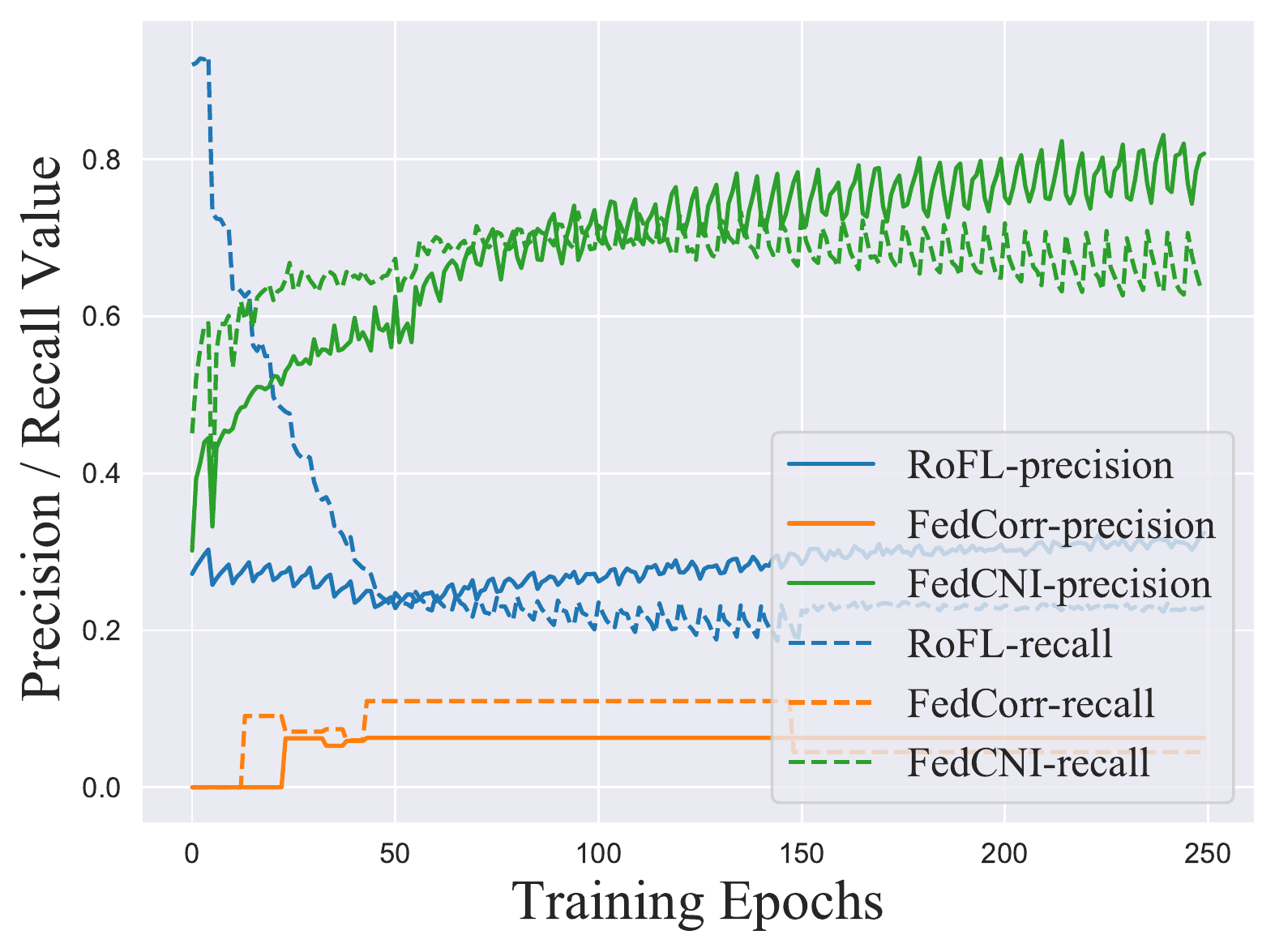}}
\subfigure[Accuracy of pseudo labeling]{\includegraphics[width=0.62\columnwidth]{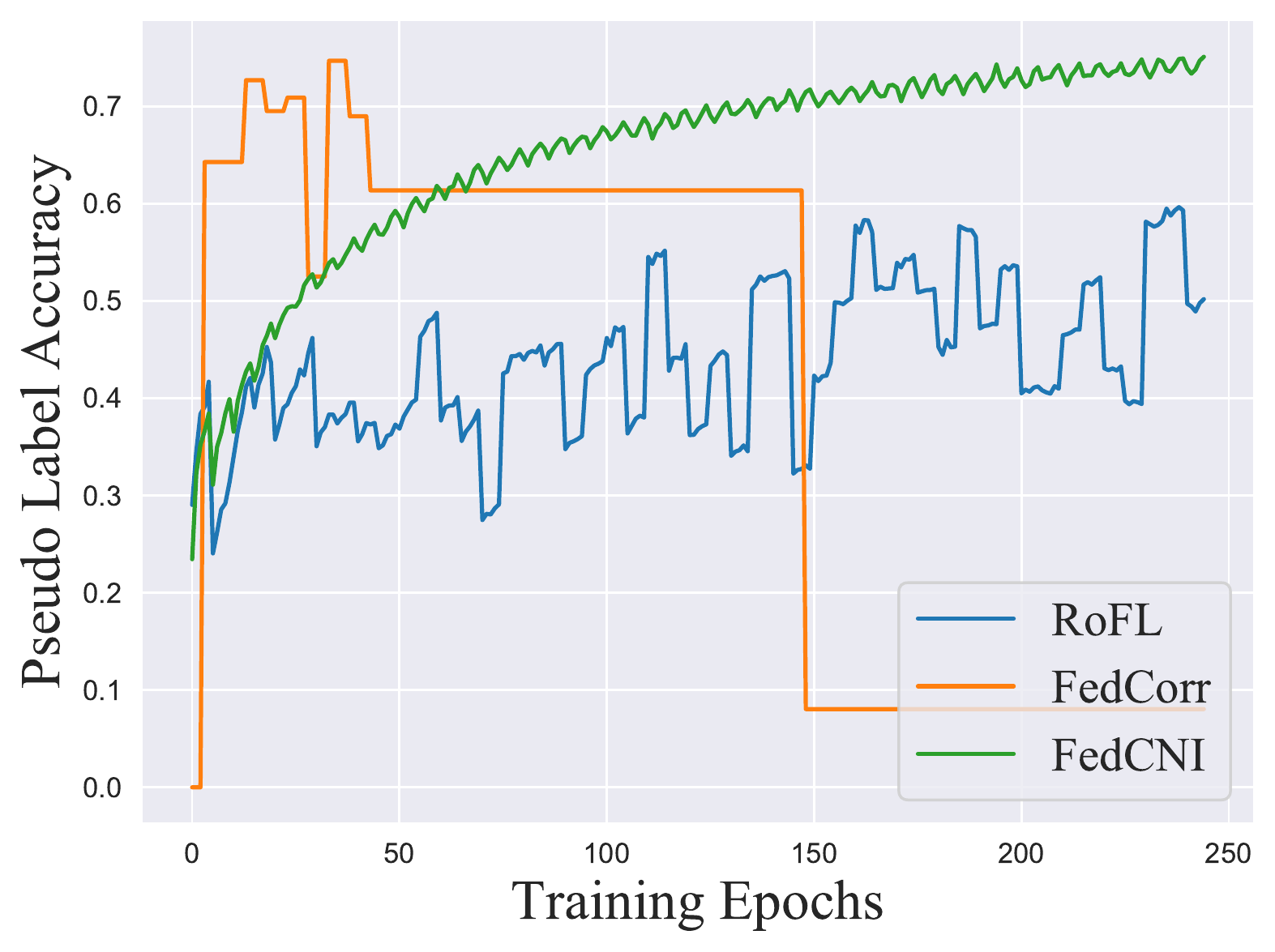}}
\vspace{-10pt}
\caption{ \textbf{Illustration of learning process.} Experiments are conducted on CIFAR-10. Our method even has comparable test accuracy convergence with clean \textsc{FedAvg} (without any noisy samples). Additionally, our method has stably higher precision \& recall of noise detection and higher accuracy of pseudo labeling compared with FL baselines with noise detection modules.}
\vspace{-10pt}
 \label{learning process} 

\end{figure*}


\subsection{Learning Process Analysis}
 
\noindent\textbf{Test Accuracy Convergence.} In Figure~\ref{learning process} (a), we illustrate the test accuracy curves in one experimental setting. \textsc{FedCNI} achieves a more stable and dominant learning curve than the baselines. Surprisingly, it even has comparable performance with clean \textsc{FedAvg} (without any noisy samples), which strongly showcases the effectiveness of \textsc{FedCNI}. 

\noindent\textbf{Noise Detection Performance.} We investigate the precision and recall of noise detection in Figure~\ref{learning process} (b). We compare our method with the baselines that have the noise detection modules, \textsc{FedCorr} and \textsc{RoFL}. We note that \textsc{FedCorr} only detects noise for limited times instead of every epoch. 
The results show that our \textsc{FedCNI} can outperform with 50\% higher than \textsc{RoFL}, 70\% higher than \textsc{FedCorr} in precision. For the recall, our method also shows a 40\% advantage over the baselines. It is notable that our noise detection stably improves along with model learning. 

\noindent\textbf{Label Correction Performance.}
To verify the effectiveness of pseudo labeling, we further observe the average accuracy between the given pseudo labels and ground-truth labels across clients. The results in Figure~\ref{learning process} (c) show that our prototypical local noise detection outperforms two noisy-label FL baselines. Our \textsc{FedCNI} can stabilize at greater than 70\% accuracy in label correction, which is 10\% to 20\% higher than baselines. \textsc{FedCorr} has high accuracy at the beginning, but it is less stable. Besides, due to the failure of noise detection, two baselines may wrongly change the clean samples' labels, leading to unsatisfactory accuracy.
\subsection{Ablation Study}
We conduct experiments to validate the effect of each component in \textsc{FedCNI}, shown in Table~\ref{ablation table}. It is found that all components help to improve the performance, showing their effectiveness. We find denoise Mixup has a greater impact than other components. 

\setlength{\tabcolsep}{2pt}
\begin{table}[t]
\vspace{-15pt}
\centering
\caption{\textbf{Ablation study.} Top-1\% test acc. on CIFAR-10.}
\resizebox{0.5\textwidth}{!}{%
\begin{tabular}{lcccc}
    \toprule
    \midrule
    Method$\backslash$(Sym./pair, $\mu,\sigma,\alpha_{Dir}$) & $(S,0.2,0.2,1)$ & $(S,0.4,0.2,0.7)$  & $(P,0.2,0.2,1)$ & $(P,0.4,0.2,0.7)$  \\
    \midrule
     \textsc{FedAvg}   & 79.94  & 63.28 
                & 80.58& 60.42  \\
     \midrule
    Ours w/o Curr. Pseudo Labeling &83.41 &  73.68 &  82.25  &65.74\\
    Ours w/o Denoise Mixup         &82.46 &  74.52 & 82.17 &67.52 \\
    Ours w/o Proto. Sim. Loss  &86.5 & 76.28  &  85.6 &71.36  \\
    Ours w/o Global Aggregator  & 84.93& 75.06  & 84.03 &68.49\\
    \midrule
     \rowcolor{gray!20}Ours    & \textbf{ 86.62} & \textbf{ 78.1} &\textbf{  86.52}&\textbf{  71.54}\\
    \bottomrule
  \end{tabular}
}
\vspace{-15pt}
\label{ablation table}  
\end{table}

\section{Conclusion}
\label{conclusion}
In this paper, we propose \textsc{FedCNI} to address the challenges brought by the co-existence of label noise and Non-IID data in FL. It includes a noise-resilient local solver and a robust global aggregator. For the local solver, we design a more robust prototypical noise detector to distinguish noisy samples. Further, to reduce the negative impact brought by the noisy samples, we devise a curriculum pseudo labeling method and a denoise Mixup training strategy. For the global aggregator, we propose a switching re-weighted aggregation method tailored to different learning periods. Extensive experiments demonstrate our method can substantially outperform state-of-the-art solutions in mix-heterogeneous FL environments. 
\bibliographystyle{IEEEtran}
\bibliography{main}

\begin{thebibliography}{10}
\providecommand{\url}[1]{#1}
\csname url@samestyle\endcsname
\providecommand{\newblock}{\relax}
\providecommand{\bibinfo}[2]{#2}
\providecommand{\BIBentrySTDinterwordspacing}{\spaceskip=0pt\relax}
\providecommand{\BIBentryALTinterwordstretchfactor}{4}
\providecommand{\BIBentryALTinterwordspacing}{\spaceskip=\fontdimen2\font plus
\BIBentryALTinterwordstretchfactor\fontdimen3\font minus
  \fontdimen4\font\relax}
\providecommand{\BIBforeignlanguage}[2]{{%
\expandafter\ifx\csname l@#1\endcsname\relax
\typeout{** WARNING: IEEEtran.bst: No hyphenation pattern has been}%
\typeout{** loaded for the language `#1'. Using the pattern for}%
\typeout{** the default language instead.}%
\else
\language=\csname l@#1\endcsname
\fi
#2}}
\providecommand{\BIBdecl}{\relax}
\BIBdecl

\bibitem{zhang2022towards}
D.~Zhang, K.~Shen, F.~Wang, D.~Wang, and J.~Liu, ``Towards joint loss and
  bitrate adaptation in realtime video streaming,'' in \emph{Proceedings of
  ICME}, 2022.

\bibitem{pmlr-v54-mcmahan17a}
B.~McMahan, E.~Moore, and et~al, ``{Communication-Efficient Learning of Deep
  Networks from Decentralized Data},'' in \emph{Proceedings of the 20th
  International Conference on Artificial Intelligence and Statistics}, 2017.

\bibitem{fedprox}
T.~Li, A.~K. Sahu, M.~Zaheer, M.~Sanjabi, A.~Talwalkar, and V.~Smith,
  ``Federated optimization in heterogeneous networks,'' in \emph{Proceedings of
  Machine Learning and Systems}, 2020.

\bibitem{feddyn}
D.~A.~E. Acar, Y.~Zhao, R.~Matas, M.~Mattina, P.~Whatmough, and V.~Saligrama,
  ``Federated learning based on dynamic regularization,'' in \emph{Proceedings
  of ICLR}, 2021.

\bibitem{9860009}
X.~Shang, Y.~Lu, Y.-M. Cheung, and H.~Wang, ``Fedic: Federated learning on
  non-iid and long-tailed data via calibrated distillation,'' in
  \emph{Proceedings of ICME}, 2022.

\bibitem{Fang_2022_CVPR}
X.~Fang and M.~Ye, ``Robust federated learning with noisy and heterogeneous
  clients,'' in \emph{Proceedings of CVPR}, 2022.

\bibitem{xu2022fedcorr}
J.~Xu, Z.~Chen, T.~Q. Quek, and K.~F.~E. Chong, ``Fedcorr: Multi-stage
  federated learning for label noise correction,'' in \emph{Proceedings of
  CVPR}, 2022.

\bibitem{9713942}
S.~Yang, H.~Park, and et~al., ``Robust federated learning with noisy labels,''
  \emph{IEEE Intelligent Systems}, vol.~37, no.~2, pp. 35--43, 2022.

\bibitem{li2019dividemix}
J.~Li, R.~Socher, and S.~C. Hoi, ``Dividemix: Learning with noisy labels as
  semi-supervised learning,'' in \emph{Proceedings of ICLR}, 2019.

\bibitem{NEURIPS2018_a19744e2}
B.~Han, Q.~Yao, X.~Yu, G.~Niu, M.~Xu, W.~Hu, I.~Tsang, and M.~Sugiyama,
  ``Co-teaching: Robust training of deep neural networks with extremely noisy
  labels,'' in \emph{Advances in Neural Information Processing Systems}, 2018.

\bibitem{9412599}
T.~Tuor, S.~Wang, B.~J. Ko, C.~Liu, and K.~K. Leung, ``Overcoming noisy and
  irrelevant data in federated learning,'' in \emph{Proceedings of ICPR}, 2021.

\bibitem{chen2020focus}
Y.~Chen, X.~Yang, X.~Qin, H.~Yu, B.~Chen, and Z.~Shen, ``Focus: Dealing with
  label quality disparity in federated learning,'' \emph{arXiv preprint
  arXiv:2001.11359}, 2020.

\bibitem{zhang2021flexmatch}
B.~Zhang, Y.~Wang, W.~Hou, H.~Wu, J.~Wang, M.~Okumura, and T.~Shinozaki,
  ``Flexmatch: Boosting semi-supervised learning with curriculum pseudo
  labeling,'' in \emph{Advances in Neural Information Processing Systems},
  2021.

\bibitem{zhang2018Mixup}
H.~Zhang, M.~Cisse, Y.~N. Dauphin, and D.~Lopez-Paz, ``mixup: Beyond empirical
  risk minimization,'' in \emph{Proceedings of ICLR}, 2018.

\bibitem{pmlr-v97-shen19e}
Y.~Shen and S.~Sanghavi, ``Learning with bad training data via iterative
  trimmed loss minimization,'' in \emph{Proceedings of International Conference
  on Machine Learning}, 2019.

\bibitem{he2016deep}
K.~He, X.~Zhang, S.~Ren, and J.~Sun, ``Deep residual learning for image
  recognition,'' in \emph{Proceedings of CVPR}, 2016.

\bibitem{krizhevsky2009learning}
A.~Krizhevsky, G.~Hinton \emph{et~al.}, ``Learning multiple layers of features
  from tiny images,'' 2009.

\bibitem{Xiao_2015_CVPR}
T.~Xiao, T.~Xia, Y.~Yang, and et~al, ``Learning from massive noisy labeled data
  for image classification,'' in \emph{Proceedings of CVPR}, 2015.

\bibitem{tan2022fedproto}
Y.~Tan, G.~Long, and et~al, ``Fedproto: Federated prototype learning across
  heterogeneous clients,'' in \emph{Proceedings of AAAI}, 2022.

\bibitem{ghosh2017robust}
A.~Ghosh, H.~Kumar, and P.~S. Sastry, ``Robust loss functions under label noise
  for deep neural networks,'' in \emph{Proceedings of AAAI}, 2017.

\bibitem{scaffold}
S.~P. Karimireddy, S.~Kale, M.~Mohri, S.~Reddi, S.~Stich, and A.~T. Suresh,
  ``{SCAFFOLD}: Stochastic controlled averaging for federated learning,'' in
  \emph{Proceedings of International Conference on Machine Learning}, 2020.

\end{thebibliography}

\end{document}